\newif\ifcmt
\cmtfalse
\documentclass[final]{cvpr}

\usepackage{times}
\usepackage{epsfig}
\usepackage{graphicx}
\usepackage{amsmath}
\usepackage{amssymb}
\usepackage{bm}
\usepackage{algorithm}
\usepackage[noend]{algpseudocode}
\usepackage{booktabs}
\usepackage{multirow}
\usepackage{xcolor}
\usepackage[normalem]{ulem}

\usepackage[pagebackref=true,breaklinks=true,colorlinks,bookmarks=false]{hyperref}

\def\cvprPaperID{5036} 
\def\confYear{CVPR 2021}
\ifcmt
\fi

\newcommand{\train}{\mathcal{D}_{\mathrm{train}}}
\newcommand{\valid}{\mathcal{D}_{\mathrm{val}}}
\newcommand{\test}{\mathcal{D}_{\mathrm{test}}}
\newcommand{\traini}{\mathcal{D}^i_{\mathrm{train}}}
\newcommand{\validi}{\mathcal{D}^i_{\mathrm{val}}}

\def\sE{{\mathbb{E}}}

\def\vone{{\bm{1}}}
\def\vtheta{{\bm{\theta}}}
\def\vlambda{{\bm{\lambda}}}
\def\vTheta{{\bm{\Theta}}}
\def\vLambda{{\bm{\Lambda}}}
\def\vmu{{\bm{\mu}}}
\def\vb{{\bm{b}}}
\def\vm{{\bm{m}}}

\def\vu{{\bm{u}}}

\def\vx{{\bm{x}}}
\def\vy{{\bm{y}}}

\def\mLambda{{\bm{\Lambda}}}
\def\mSigma{{\bm{\Sigma}}}

\def\mH{{\bm{H}}}
\def\mI{{\bm{I}}}

\DeclareMathAlphabet{\mathsfit}{\encodingdefault}{\sfdefault}{m}{sl}
\SetMathAlphabet{\mathsfit}{bold}{\encodingdefault}{\sfdefault}{bx}{n}

\def\sM{{\mathbb{M}}}
\def\sN{{\mathbb{N}}}

\DeclareMathOperator*{\argmin}{arg\,min}

\begin{document}

\title{AutoDO: Robust AutoAugment for Biased Data with Label Noise\\via Scalable Probabilistic Implicit Differentiation}

\author{Denis~Gudovskiy\textsuperscript{\rm 1}
\qquad~Luca~Rigazio\textsuperscript{\rm 2}\\
\qquad~Shun~Ishizaka\textsuperscript{\rm 3}
\qquad~Kazuki~Kozuka\textsuperscript{\rm 3}
\qquad~Sotaro~Tsukizawa\textsuperscript{\rm 3} \\
{\textsuperscript{\rm 1} Panasonic AI Lab, USA} ~~
{\textsuperscript{\rm 2} AIoli Labs, USA} ~~
{\textsuperscript{\rm 3} Panasonic Technology Division, Japan} \\
\small{\texttt{denis.gudovskiy@us.panasonic.com}}
\qquad\small{\texttt{luca@aiolilabs.com}}\\
\small{\texttt{\{ishizaka.shun, kozuka.kazuki, tsukizawa.sotaro\}@jp.panasonic.com}}
}

\maketitle
\ifcmt
  \thispagestyle{empty}
\fi
\begin{abstract}
AutoAugment~\cite{cubuk2018autoaugment} has sparked an interest in automated augmentation methods for deep learning models. These methods estimate image transformation policies for train data that improve generalization to test data. While recent papers evolved in the direction of decreasing policy search complexity, we show that those methods are not robust when applied to biased and noisy data. To overcome these limitations, we reformulate AutoAugment as a generalized automated dataset optimization (AutoDO) task that minimizes the distribution shift between test data and distorted train dataset. In our AutoDO model, we explicitly estimate a set of per-point hyperparameters to flexibly change distribution of train data. In particular, we include hyperparameters for augmentation, loss weights, and soft-labels that are jointly estimated using implicit differentiation. We develop a theoretical probabilistic interpretation of this framework using Fisher information and show that its complexity scales linearly with the dataset size. Our experiments on SVHN, CIFAR-10/100, and ImageNet classification show up to 9.3\% improvement for biased datasets with label noise compared to prior methods and, importantly, up to 36.6\% gain for underrepresented SVHN classes\footnote{\href{https://github.com/gudovskiy/autodo}{Our code is available at github.com/gudovskiy/autodo}}.
\end{abstract}

\section{Introduction}
\label{sec:intro}
Data augmentation (DA) plays one of the key roles in improving accuracy of deep neural networks (DNNs)~\cite{alexnet}. DA increases size and diversity of train dataset and consequently improves generalization to test data distribution. Unfortunately, DA design requires expert domain knowledge, dataset analysis, and numerous costly experiments. In real applications with biased~\cite{terhorst2021comprehensive} and noisy-label data~\cite{zhangoverfit}, the handpicking of DA becomes a challenging task.

\begin{figure}[t]
	\centering
	\includegraphics[width=0.98\columnwidth]{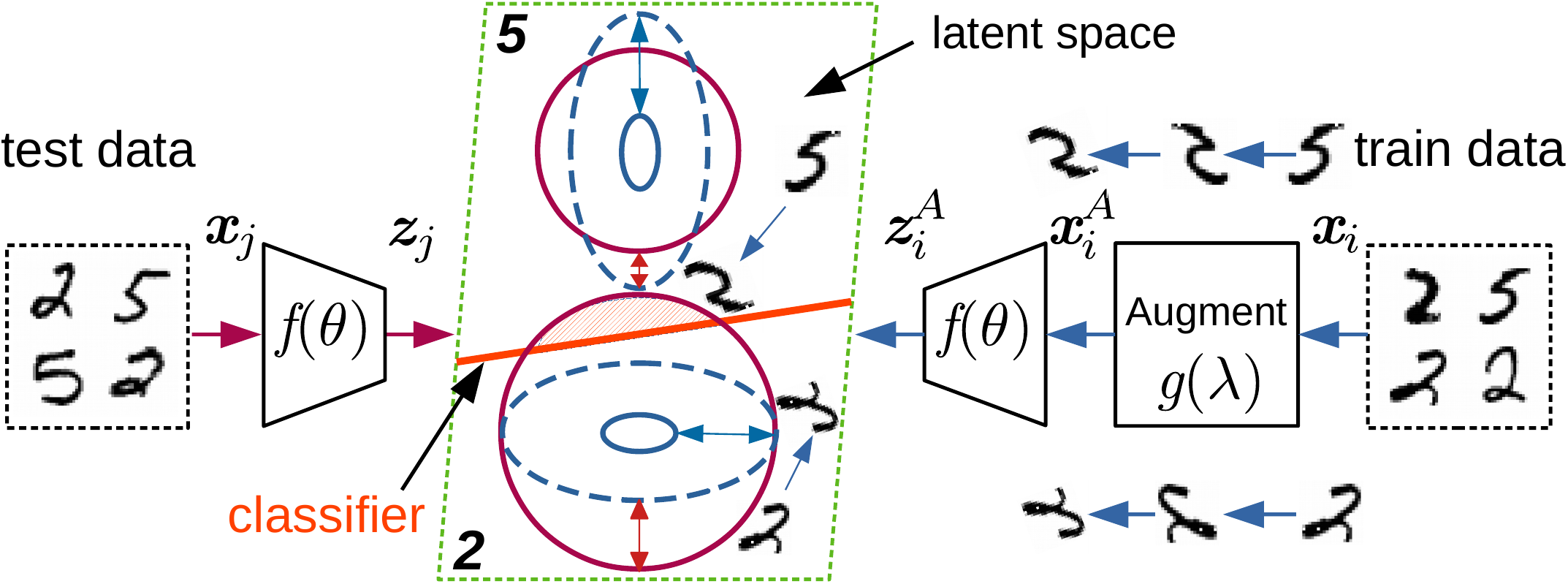}
	\caption{Shared-policy DA dilemma: the distribution of augmented train data (dashed blue) may not match the test data (solid red) in the latent space: "2" is under-augmented, while "5" is over-augmented. As a result, prior methods cannot match the test distribution and the decision of the learned classifier $f(\vtheta)$ is inaccurate.}
	\label{fig:prob}
\end{figure}

The automation of DA aims to estimate data transformation models without incurring the aforementioned difficulties. The seminal AutoAugment~\cite{cubuk2018autoaugment} (AA) proposes a DA policy model estimated by reinforcement learning (RL). Though AA outperforms prior classification baselines, its policy search takes thousands of GPU hours. Follow-up works~\cite{ho2019pba, lim2019fast, hataya2019faster} address the search complexity problem while keeping comparable accuracy results.

\begin{figure*}[ht]
	\centering
	\includegraphics[width=0.92\textwidth]{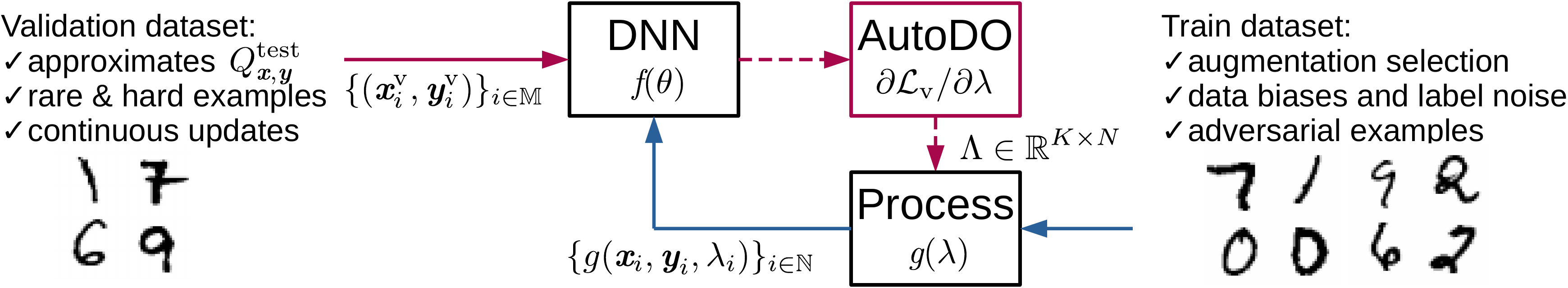}
	\caption{General setting: the validation dataset is selected to approximate the test data distribution $Q^{\mathrm{test}}_{\vx,\vy}$. The train dataset is subject to the specified data distortions and unknown DA hyperparameters. Unlike prior DA models with shared policies for all train data points, our AutoDO model $g(\vlambda)$ estimates hyperparameters $\vlambda_{i=1 \ldots N} \in \mathbb{R}^{K \times 1}$ for each of $N$ train data points via implicit differentiation framework.}
	\label{fig:general}
\end{figure*}

In this paper, we show that existing methods are not robust to train dataset distortions such as distribution bias and noisy labels. We attribute this to the policy model that \textit{shares parameters} across all data points. We illustrate this with the toy classification task in Figure~\ref{fig:prob}. The distribution of digits "2" and "5" in the train dataset (solid blue) is uneven and less diverse than test distribution (solid red) in the latent space before a linear classifier. Then, the estimated augmentation policies virtually increase train data size (dashed blue line). However, such DA model \textit{evenly} increases diversity of all digits: digit "2" is under-augmented, while digit "5" is over-augmented. This pushes decision boundary of the linear classifier in the wrong direction. A multi-class classifier is even less robust~\cite{terhorst2021comprehensive}, specifically, when it overfits to examples with noisy labels~\cite{zhangoverfit}.

To overcome these limitations, we propose to estimate DA hyperparameters for \textit{each train data point}. Moreover, our model includes and jointly optimizes loss weights to capture data biases and soft-labels to address noisy labels. This reformulates the original AA shared-policy search task into a generalized automated dataset optimization (AutoDO). The objective of our AutoDO model is to match the distribution of a small clean unbiased validation dataset with a large distorted train dataset using a set of per-point hyperparameters as illustrated in Figure~\ref{fig:general}. We optimize our model with large-scale hyperparameters using implicit differentiation~\cite{lorraine-million} and analytically show that it is equivalent to maximizing the Fisher information between empirical datasets. Experiments on class-imbalanced data with noisy labels show the advantages of our approach.

\section{Related work}
\label{sec:related}
DA has been widely used in computer vision from shallow convolutional neural networks (CNNs) for MNIST~\cite{simard2003best} to deep CNNs for large-scale ImageNet~\cite{alexnet} recognition. Other applications such as segmentation~\cite{Chen2018DeepLabSI} and object detection~\cite{ssd} benefit from DA as well. Recent Mixup~\cite{zhang2018mixup} and CutMix~\cite{yun2019cutmix} heuristically mix images to produce better regularization at the semantic level. However, all these methods require to handpick a set of hyperparameters.

Naturally, the idea to estimate DA hyperparameters emerged in AutoAugment~\cite{cubuk2018autoaugment} (AA) by sequentially sampling policies and using validation accuracy as a reward to update the RL controller. However, AA requires thousands of GPU hours to achieve results superior to the baseline. To speed up policy search, Ho~\etal~\cite{ho2019pba} propose population-based training~\cite{jaderberg2017population} that can be efficiently parallelized across distributed CPUs or GPUs. Unlike AA, this population-based augmentation (PBA) searches for DA policy schedules rather than individual augmentations. Recent Fast AutoAugment~\cite{lim2019fast} (FAA) further extends this idea by searching for policy schedules that maximize the match between distribution of augmented train data and unaugmented validation data. Inspired by Tran~\etal~\cite{NIPS2017_6872}, Bayesian optimization is adopted by FAA to achieve density matching.

Though the methods in~\cite{ho2019pba,lim2019fast} substantially decrease the complexity of AA~\cite{cubuk2018autoaugment}, gradient-based optimization can be scaled up to even larger models and datasets. Furthermore, it is directly supported by differentiable DNNs and ML frameworks. Zhang~\etal~\cite{zhang2020adversarial} and Lin~\etal~\cite{Lin_2019_ICCV} adopt the REINFORCE gradient estimator~\cite{Williams92simplestatistical}, which further decreases learning time. Authors use the same policy model of~\cite{cubuk2018autoaugment, lim2019fast}, but estimate DA hyperparameters online within the bilevel gradient optimization framework. In this framework, an inner objective minimizes conventional train loss, while an outer objective minimizes adversarial~\cite{zhang2020adversarial} or validation~\cite{Lin_2019_ICCV} loss. Similarly, Hataya~\etal~\cite{hataya2019faster} and Li~\etal~\cite{li2020dada} employ the DARTS gradient estimator~\cite{liu2018darts} inspired by its success in a neural architecture search.

Lastly, RandAugment~\cite{cubuk2019randaugment} and UniformAugment~\cite{lingchen2020uniformaugment} propose to dramatically decrease policy search space by a reparameterization scheme and to manually select only few DA hyperparameters. These near search-free methods are almost as effective as the computationally expensive search-based methods~\cite{cubuk2018autoaugment,lim2019fast}. This observation raises questions on the \textit{applicability and optimality} of the search-based methods. In this paper, we show that these DA methods with the same shared-policy model perform very similarly only on undistorted train datasets. At the same time, the search-based methods can be significantly more effective and robust in the case of biased datasets with label noise. However, prior search-based methods with the shared-policy model are ill-equipped to handle distorted datasets that is sketched in Figure~\ref{fig:prob}. Motivated by this, we develop our AutoDO model with the following contributions:
\begin{itemize}
	\itemsep0em
	\item To control train dataset distribution more accurately, we introduce the AutoDO model that jointly optimizes per-point hyperparameters for DA, loss weights and soft-labels using implicit differentiation.
	\item We analytically show that implicit differentiation minimizes the distribution shift between validation and train data by maximizing Fisher information. We also show a connection to DARTS~\cite{liu2018darts} used in~\cite{hataya2019faster,li2020dada}.
	\item Our model outperforms prior methods in image classification by up to 9.3\% when the task model is trained on class-imbalanced data with label noise. More importantly, we show that it improves accuracy of underrepresented classes: by up to 36.6\% for SVHN.
\end{itemize}

\section{Theoretical background}
\label{sec:theory}
\subsection{Problem statement}
\label{subsec:problem}

Let $(\vx,\vy)$ be an input-label pair where a given label $\vy=\vone_c\in\mathbb{B}^{C \times 1}$ is one-hot vector (\textit{hard-label}) for classification task with $C$ classes. A validation dataset $\valid=\{(\vx^\mathrm{v}_i,\vy^\mathrm{v}_i)\}_{i\in\sM}$ of size $M$ approximates test data $\test$. A train dataset $\train=\{(\vx_i,\vy_i)\}_{i\in\sN}$ of size $N$ may have a different from $\valid$ distribution and noisy labels $\vy$.

There are a vector of parameters $\vtheta$ for a \textit{task model} $f(\vtheta)$ and a vector $\vlambda$ of \textit{hyperparameters}. The latter only \textit{implicitly} influences the train loss $\mathcal{L}(\vlambda, \vtheta)$ such that the optimal parameters $\vtheta^{\ast}$ can be written as an \textit{implicit function} of hyperparameters $\vtheta^{\ast}(\vlambda)$. Then, the \textit{inner objective} optimization can be expressed as
\begin{equation} \label{eq:bi1}
	\vtheta^{\ast}(\vlambda) := \argmin_{\vtheta} \mathcal{L}(\vlambda, \vtheta),
\end{equation}
where the train loss is the empirical risk: $\mathcal{L}(\vlambda, \vtheta) = \sum\nolimits_{i \in \sN} \mathcal{L}(\vy_i,\hat{\vy}_i) / N = \sum_{i \in \sN} \mathcal{L}(\vy_i,f(\vx_i, \vtheta(\vlambda))) / N$.

Since test distribution $Q^{\mathrm{test}}_{\vx,\vy}$ is not known, it is usually replaced by an empirical distribution $\hat{Q}^{\mathrm{val}}_{\vx,\vy}$ of the validation dataset $\valid$~\cite{lim2019fast}. Then, the optimal hyperparameters $\vlambda^{\ast}$ minimize an \textit{outer objective} as
\begin{equation} \label{eq:bi2}
\vlambda^{\ast} := \argmin_{\vlambda} \mathcal{L}^{\ast}_\mathrm{v}(\vlambda) = \argmin_{\vlambda} \mathcal{L}_\mathrm{v}(\vlambda, \vtheta^{\ast}(\vlambda)),
\end{equation}
where the validation dataset loss $\mathcal{L}_\mathrm{v}(\vlambda, \vtheta^{\ast}(\vlambda)) = \sum\nolimits_{i \in \sM} \mathcal{L}(\vy^\mathrm{v}_i,\hat{\vy}^\mathrm{v}_i) / M = \sum_{i \in \sM} \mathcal{L}(\vy^\mathrm{v}_i,f(\vx^\mathrm{v}_i, \vtheta^{\ast}(\vlambda))) / M$, and it does not \textit{explicitly} depend on $\vlambda$.

A solution for~(\ref{eq:bi1}-\ref{eq:bi2}) problem statement can be found by \textit{bilevel} (nested) optimization procedure~\cite{Lin_2019_ICCV,li2020dada,hataya2019faster}, where the inner objective is the conventional training to find $\vtheta^{\ast}$ and the outer objective is a hyperparameter optimization (HO) to estimate $\vlambda^{\ast}$. Unlike search-free methods~\cite{cubuk2019randaugment,lingchen2020uniformaugment}, HO can find nearly optimal $\vlambda^{\ast}$ in automated fashion. However, the solution to bilevel optimization can be computationally costly for large-scale hyperparameters. We find a scalable solution using gradient-based HO for fully-differentiable DNN models. Specifically, our AutoDO model relies on an implicit differentiation framework.

\subsection{Solution using implicit differentiation}
\label{subsec:gradient}

Since differentiation of (\ref{eq:bi1}) is known~\cite{paszke2017automatic}, we are only interested in obtaining $\partial \mathcal{L}_\mathrm{v}/\partial \vlambda$ in (\ref{eq:bi2}). This derivative at a point $(\vlambda, \vtheta^{\ast}(\vlambda))$ can be found using a chain rule as
\begin{equation} \label{eq:bi3}
\frac{\partial \mathcal{L}_\mathrm{v}}{\partial \vlambda} = \frac{\partial \mathcal{L}_\mathrm{v}}{\partial \vlambda} + \frac{\partial \mathcal{L}_\mathrm{v}}{\partial \vtheta^{\ast}(\vlambda)} \frac{\partial \vtheta^{\ast}(\vlambda)}{\partial \vlambda} = \frac{\partial \mathcal{L}_\mathrm{v}}{\partial \vtheta^{\ast}(\vlambda)} \frac{\partial \vtheta^{\ast}(\vlambda)}{\partial \vlambda},
\end{equation}
where the direct derivative $\partial \mathcal{L}_\mathrm{v}/\partial \vlambda=0$, because $\mathcal{L}_\mathrm{v}$ by definition does not explicitly depend on $\vlambda$. The derivative $\partial \mathcal{L}_\mathrm{v}/\partial \vtheta^{\ast}(\vlambda)$ can be computed using conventional differentiation. However, the derivative $\partial \vtheta^{\ast}(\vlambda)/\partial \vlambda$ is not known because implicit function $\vtheta^{\ast}(\vlambda)$ is not defined.

Recently, Lorraine~\etal~\cite{lorraine-million} proposed to use the implicit function theorem (IFT)~\cite{cauchy} to find $\partial \vtheta^{\ast}(\vlambda)/\partial \vlambda$. Let $S(\vlambda, \vtheta)=\partial \mathcal{L}(\vlambda, \vtheta)/\partial \vtheta : \vLambda \times \vTheta \rightarrow \vTheta$ be a continuously differentiable function ($C^1$). For a fixed point $(\acute{\vlambda},\acute{\vtheta})$, if $S(\acute{\vlambda},\acute{\vtheta})=0$ and $\det J_{\vtheta} S(\acute{\vlambda},\acute{\vtheta})\neq0$: a) there exists an implicit function $\vtheta = s(\vlambda)$ for $\| \vlambda - \acute{\vlambda} \| \le r_1$ and $\| \vtheta - \acute{\vtheta} \| \le r_2$, b) the function $\vtheta = s(\vlambda)$ is of class $C^1$, and its Jacobian is defined as
\begin{equation} \label{eq:ift1}
\partial \vtheta(\vlambda)/\partial \vlambda = -\left[J_{\vtheta} S(\vlambda, \vtheta)\right]^{-1} J_{\vlambda} S(\vlambda, \vtheta).
\end{equation}

Using the definition of $S(\vlambda, \vtheta)$ and (\ref{eq:ift1}), we rewrite~(\ref{eq:bi3}) as
\begin{equation} \label{eq:ift2}
\frac{\partial \mathcal{L}_\mathrm{v}}{\partial \vlambda} = 
-\frac{\partial \mathcal{L}_\mathrm{v}}{\partial \vtheta} \left[\frac{\partial^2 \mathcal{L}}{\partial \vtheta \partial \vtheta^T}\right]^{-1} \frac{\partial^2 \mathcal{L}}{\partial \vtheta \partial \vlambda^T}.
\end{equation}

The difficulty in implicit differentiation~(\ref{eq:ift2}) is to find inverse of Hessian $\mH_{\vtheta}^{-1} = \left[\partial^2 \mathcal{L}/(\partial \vtheta \partial \vtheta^T)\right]^{-1}$, which scales quadratically $(\mH_{\vtheta}\in\mathbb{R}^{L \times L})$ with the size of task model parameters $\vtheta\in\mathbb{R}^{L \times 1}$. Fortunately, ML frameworks~\cite{paszke2017automatic} are able to calculate Jacobian-vector products to reduce memory. Computationally, $\mH_{\vtheta}^{-1}$ can be approximated by conjugate gradients~\cite{pedregosa2016hyperparameter,rajeswaran2019meta}, quasi-Newton methods~\cite{deq}, or Neumann series~\cite{if,lorraine-million}. We rely on the latter approach.

\begin{figure*}[ht]
	\centering
	\includegraphics[width=0.7\textwidth]{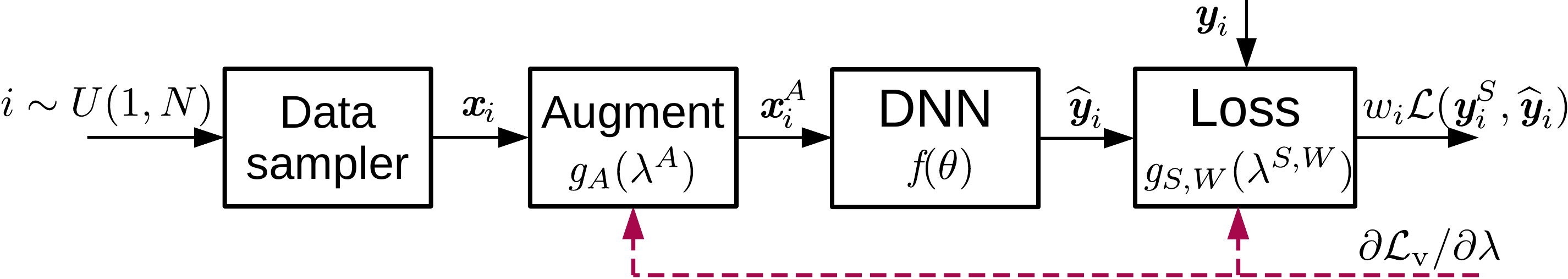}
	\caption{Data pipeline of AutoDO model $g(\vlambda)$ for joint dataset optimization: augmentation, loss reweighting and soft-labeling are accomplished by, correspondingly, $g_A(\vlambda^A)$, $g_W(\vlambda^W)$ and $g_S(\vlambda^S)$ using per-point hyperparameter vectors $\vlambda=[\vlambda^A; \vlambda^W; \vlambda^S]\in\mathbb{R}^{K \times 1}$.}
	\label{fig:augment}
\end{figure*}

\begin{figure}[ht]
	\centering
	\includegraphics[width=0.7\columnwidth]{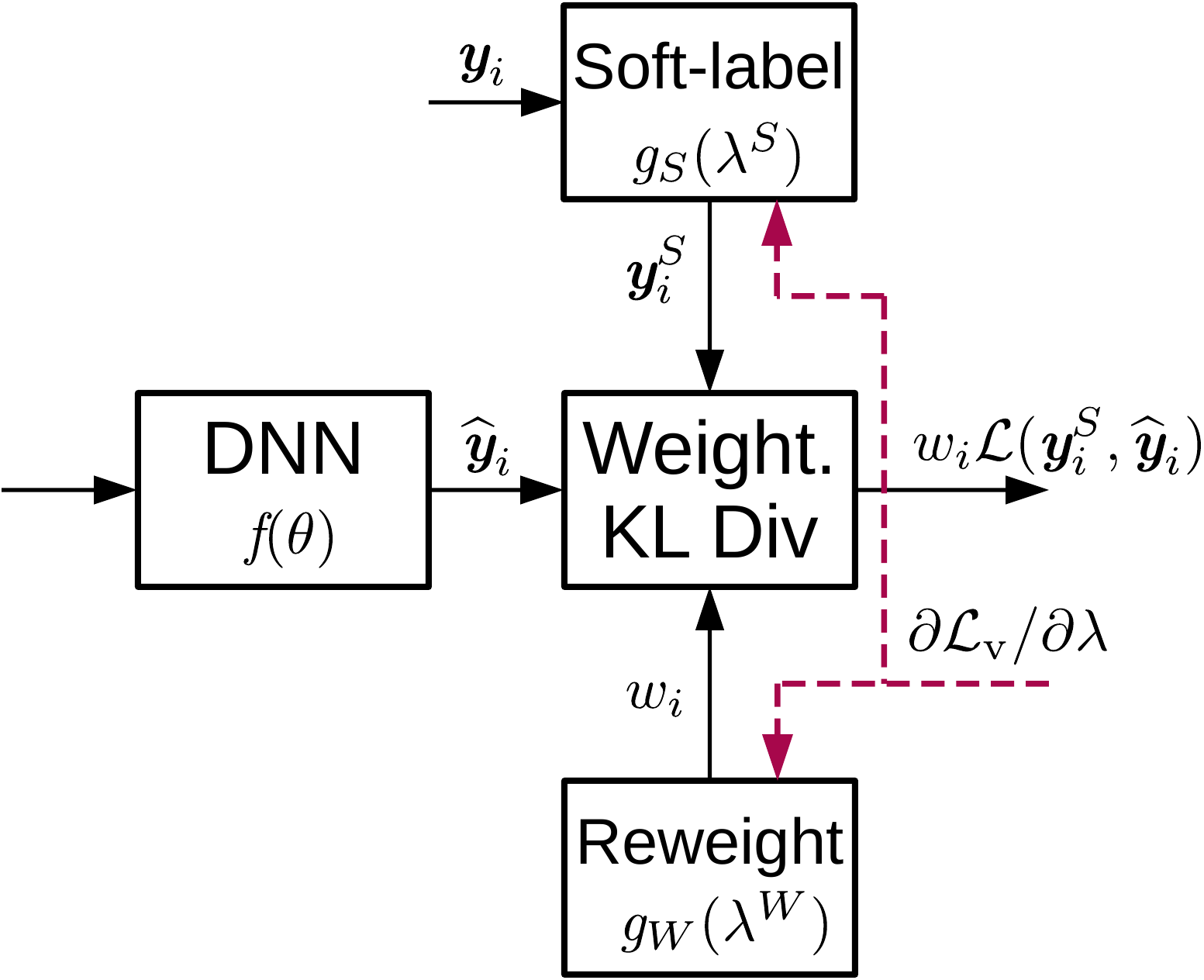}
	\caption{Diagram of the proposed AutoDO for loss reweighting and soft-labeling: conventional cross-entropy loss for classification is replaced by a weighted Kullback-Leibler (KL) divergence.}
	\label{fig:loss}
\end{figure}

\subsection{Implicit differentiation meets density matching}
\label{subsec:density}

One may ask what is the meaning of implicit differentiation~(\ref{eq:ift2}) in the context of the bilevel framework~(\ref{eq:bi1}-\ref{eq:bi2})? The classification loss in~(\ref{eq:bi1}) typically minimizes Kullback-Leibler (KL) divergence between joint train data distribution $Q_{\vx,\vy}$ with density $q(\vx,\vy)$ and the learned model distribution $P_{\vx,\vy}(\vtheta)$ with $p(\vx,\vy | \vtheta)$ density. However, it can be shown that KL objective learns only a \textit{conditional distribution} $p(\vy | \vx, \vtheta)$ of the train data distribution $Q_{\vy|\vx}$~\cite{gudovskiy2020al}. In other words, $f(\vtheta)$ is a strictly \textit{discriminative model}.

The KL loss~(\ref{eq:bi1}) for the empirical train distribution $\hat{Q}_{\vx}$ with hard-labels $\vy$ can be simplified to
\begin{equation} \label{eq:meet1}
\mathcal{L}(\vlambda, \vtheta) = -\sum\nolimits_{i \in \sN} \log p(\vy_i|\vx_i, \vtheta(\vlambda))/N.
\end{equation}

By substituting~(\ref{eq:meet1}) into each term in~(\ref{eq:ift2}) and rewriting them as gradients for $i$th data point, it can be shown that $\vu^\mathrm{v}_j(\vtheta)=-\partial \mathcal{L}_\mathrm{v}(j)/\partial \vtheta = \nabla_{\vtheta}\log p(\vy^\mathrm{v}_j|\vx^\mathrm{v}_j, \vtheta(\vlambda))$ and $\vu_i(\vtheta)=-\partial \mathcal{L}(i)/\partial \vtheta=\nabla_{\vtheta}\log p(\vy_i|\vx_i, \vtheta(\vlambda))$ are the Fisher scores \wrt validation and train data points~\cite{learningfisher}, respectively. Moreover, the expectation of Hessian $-\sE_{\hat{Q}_{\vx}} \left[ \mH_{\vtheta} \right]$ is equal to Fisher information metric $\mathcal{\mI}_{\vtheta}=\sum\nolimits_{i \in \sN} \vu_i(\vtheta) \vu_i(\vtheta)^T / N$~\cite{lyfisher}.

Then, the gradient of (\ref{eq:ift2}) for empirical datasets simply maximizes Fisher kernel between $\hat{Q}^\mathrm{val}_{\vx}$ and $\hat{Q}_{\vx}$ as
\begin{equation} \label{eq:meet3}
\nabla_{\vlambda} \sE_{\hat{Q}^\mathrm{val}_{\vx}, \hat{Q}_{\vx}} \left[ \mathcal{L}_\mathrm{v} \right] = \sE_{\hat{Q}^\mathrm{val}_{\vx}} \left[ \vu^\mathrm{v}(\vtheta) \right] \mathcal{\mI}_{\vtheta}^{-1} \sE_{\hat{Q}_{\vx}} \left[ \vu(\vtheta) \vu(\vlambda)^T \right],
\end{equation}
\ifcmt
  where the detailed derivations are presented in Appendix~C.
\else
  where the detailed derivations are presented in Appendix~\ref{sec:supp_density}.
\fi

Therefore, the implicit differentiation in~(\ref{eq:ift2}) finds the gradient~(\ref{eq:meet3}) for an outer objective~(\ref{eq:bi2}) that changes hyperparameters $\vlambda$ in the direction to match density of $\valid$ by $\train$ density on a Riemannian manifold with a local metric given by Fisher information~\cite{domingos2020model}. If replace $\mathcal{\mI}_{\vtheta}$ with the identity matrix and recall the IFT condition $\partial \mathcal{L}(\acute{\vlambda}, \acute{\vtheta})/\partial \acute{\vtheta}=0$, (\ref{eq:meet3}) is equivalent to the DARTS gradient estimator~\cite{liu2018darts} used in prior gradient-based methods~\cite{hataya2019faster,li2020dada}.

Typically, we can assume a setup in Figure~\ref{fig:general} with a \textit{hybrid} model where the discriminative model $f(\vtheta)$ is separated from an additional model $g(\vlambda)$. The latter changes the data distribution $p(\vx,\vy | \vlambda)$. Hence, $g(\vlambda)$ can be viewed as a \textit{generative model} with the outer objective~(\ref{eq:bi2}) to minimize KL divergence between $\hat{Q}^{\mathrm{val}}_{\vx,\vy}$ and $\hat{Q}_{\vx,\vy}$ distributions~\cite{jakkola}.

\section{The proposed AutoDO model}
\label{sec:model}

We use the theoretical framework introduced in Section~\ref{sec:theory} to \textit{jointly optimize} $\train$ hyperparameters. Figure~\ref{fig:general} shows a general data setup: $\valid$ is selected to approximate $\test$ distribution, and $\train$ is subject to data distortions and unknown hyperparameters. In addition to a task model $f(\vtheta)$, there is the proposed AutoDO model $g(\vlambda)$. The latter is parameterized by a vector $\vlambda$ and can be applied to input or output of $f(\vtheta)$. In contrast to prior models, we propose to use per-point dataset hyperparameters $\vlambda_i\in \mathbb{R}^{K \times 1}$ that form a matrix $\mLambda \in \mathbb{R}^{K \times N}$ for dataset $\train$ of size $N$.

Figure~\ref{fig:augment} shows a diagram of our pipeline, where the AutoDO model $g(\vlambda)$ is divided into application-specific sub-models: $g_A(\vlambda^A)$ for augmentation, $g_W(\vlambda^W)$ for loss reweighting, and $g_S(\vlambda^S)$ for soft-labeling. Hence, a generic vector $\vlambda_i$ is a concatenation of application-specific vectors: $\vlambda_i\in\mathbb{R}^{K \times 1}=[\vlambda_i^A; \vlambda_i^W; \vlambda_i^S]\in\mathbb{R}^{(A+W+S) \times 1}$.

A conventional data sampler uniformly generates train data points $\vx_i$ that are passed to the augmentation sub-model to perform $g_{A}(\vx_i, \vlambda^A_i)$ in Figure~\ref{fig:augment}. This augmentation block outputs a modified train data point $\vx^A_i$ to the task model for training procedure. Simultaneously, soft-labels and loss weights are propagated to calculate a weighted Kullback-Leibler (KL) divergence loss $w_i \mathcal{L}(\vy^S_i, \hat{\vy}_i)$. This part is detailed in Figure~\ref{fig:loss}.

\subsection{Augmentation sub-model}
\label{subsec:augment}

Our augmentation sub-model contains binary probabilities $\vb_i\in\mathbb{B}^{A \times 1}$ and continuous magnitudes $\vm_i\in\mathbb{R}^{A \times 1}$ that parameterize $A$ operations for each $i$th data point. We argue that it is less restrictive to model each data point distribution compared to finite-length policies in~\cite{cubuk2018autoaugment, lim2019fast}.

We use a multivariate Gaussian prior for magnitudes: $\vm_i \sim \mathrm{rng} \frac{M}{10} \mathcal{N}\left(\vmu, \mSigma_i \right)$, where $M$ normalizes magnitudes as in~\cite{cubuk2019randaugment}. The mean vector $\vmu\in\mathbb{R}^{A \times 1}$ is usually consists of zeros or ones depending on operation. Only the covariance matrix $\mSigma_i\in\mathbb{R}^{A \times A}$ is parameterized by AutoDO. Although it is possible to model multivariate covariance at the expense of quadratically-growing hyperparameters, we experiment with an univariate option: $\mSigma_i=\sigma (\textrm{diag} (\vlambda^{A_m}_i))$, where $\vlambda^{A_m}_i\in\mathbb{R}^{A \times 1}$ is the magnitude hyperparameter vector and $\sigma$ is a conventional sigmoid nonlinearity. We initialize the magnitude hyperparameters to zeros, which sets the variances of $\mSigma_i$ to be in the middle of available range ($\mathrm{rng}$).

While the widely-used reparameterization trick allows to obtain gradients for magnitudes $\vm_i$ with continuous Gaussian distribution, it is not trivial for the binomial probability distribution of each operation: $\vb_i \sim \mathrm{Bern} (\sigma (\vlambda^{A_b}_i)) \in \{0,1\}$. We use Gumbel-softmax~\cite{jang2016categorical} to relax backpropagation through the discrete Bernoulli distribution. Initially, we set probability of enabling the $a$th operation $b^{a}_i$ to 25\%. 

Finally, we model a sequence ($a = 1 \ldots A$) of augmentation operations $\mathcal{O}: \mathcal{X} \rightarrow \mathcal{X}$ on the image space $\mathcal{X}$ as
\begin{equation} \label{eq:aug1}
\vx^{A}_i(a) = 
    \begin{cases}
	\mathcal{O}(\vx^{A}_i(a-1), m^{a}_i), & \text{if $b^{a}_i = 1$}\\
	\vx^{A}_i(a-1), & \text{otherwise},
	\end{cases}
\end{equation}
where $\vx^{A}_i(0) = \vx_i$ is the input to augmentation sub-model.

The number of hyperparameters in this particular sub-model $\vlambda^A_i = [\vlambda^{A_m}_i; \vlambda^{A_b}_i] \in\mathbb{R}^{2A \times 1}$ scales linearly with the number of augmentations $A$ and train data $i\in\sN$. In general, it is possible to jointly model augmentations not only along $A$-dimension using $\mSigma_i$, but also along sample dimension using \eg Gaussian mixture model.

\subsection{Reweighting and soft-labeling sub-models}
\label{subsec:loss}

Our loss reweighting sub-model multiplies $i$th train loss by a scalar: $g_W(\mathcal{L}_i, \lambda^W_i) = w_i \mathcal{L}_i$, where the weights are parameterized as: $w_i = 1.44 \times \mathrm{softplus} (\lambda^W_i)$. Hyperparameters $\lambda^W_i$ are initialized to zeros such that the weights $w_i$ are ones at the start and their output is in range $[0:\infty)$.

Although soft-label estimation in the presence of noise has been recently studied in~\cite{tanaka2018joint, Yi_2019_CVPR}, we approach it as part of joint optimization procedure with other types of applications. Moreover, we apply implicit differentiation rather than the alternating optimization as in~\cite{tanaka2018joint, Yi_2019_CVPR}.

Our soft-label sub-model consists of hyperparameters $\vlambda^S_i\in\mathbb{R}^{C \times 1}$ that are estimated for each label in $\train$ from the noise-free $\valid$. Similarly to~\cite{Yi_2019_CVPR}, we set soft-labels as follows: $\vy^S_i = g_S(\vy_i, \vlambda^S_i) = \mathrm{softmax}(\vlambda^S_i)$. But, unlike~\cite{Yi_2019_CVPR}, we initialize $\vlambda^S_i$ at epoch $e=0$ in a way to output smooth-labels~\cite{Szegedy2016RethinkingTI}: $\vy^S_i(0) = (1-\alpha)\vy_i + \alpha/C$. Our hyperparameters at the initialization step are: $\vlambda^S_i(0) = (\vy_i - 0.5)\log(1-C-C/\alpha)$, where $\vy_i = \vone_c\in\mathbb{B}^{C \times 1}$ are potentially noisy hard-labels and $\alpha$ is a small constant. The $\alpha$ constant is usually in the range $\alpha=[0.05-0.2]$~\cite{Szegedy2016RethinkingTI}.

Lastly, we replace the conventional asymmetric KL loss in~(\ref{eq:bi1}) with its symmetric version. This step is necessary because we simultaneously optimize conditional distribution $p(\vy | \vx, \vtheta)$ and joint data distribution $p(\vx,\vy | \vlambda)$.

\begin{algorithm}[t]
	\caption{AutoDO bilevel optimization of $f(\vtheta)$, $g(\vlambda)$}
	\label{alg:1}
	\begin{algorithmic}[1]
		\State Initialize parameters $\vtheta$ and hyperparameters $\vlambda$
		\For{$\mathrm{epoch} = 1 \dots \mathrm{epochs}$}
		\For{$\mathrm{batch} = 1 \dots \mathrm{batches}$}
		\State sample batch $\{(\vx,\vy)\}_{b_i\in\mathcal{B}}$ from $\train$
		\State augment data $\vx^A = g_A(\vx,\vlambda^A)$
		\State predict $\hat{\vy} = f(\vx^A,\vtheta)$
		\State generate soft-labels $\vy^S = g_S(\vy,\vlambda^{S})$
		\State calculate $\nabla_{\vtheta} \left(w \mathcal{L}(\vy^S,\hat{\vy})\right)$
		\State update $\vtheta$ using task optimizer
		\EndFor
		\If{$\mathrm{epoch}>E$}
		\For{$\mathrm{batch} = 1 \dots \mathrm{batches}$}
		\State sample $\{(\vx,\vy)\}_{b_i\in\mathcal{B}}$ from $\train$
		\State sample $\{(\vx^\mathrm{v},\vy^\mathrm{v})\}_{b_j\in\mathcal{B}}$ from $\valid$
		\State predict $\hat{\vy}^\mathrm{v} = f(\vx^\mathrm{v},\vtheta)$
		\State predict $\hat{\vy} = f(\vx^A,\vtheta) = f(g_A(\vx,\vlambda^A),\vtheta)$
		\State generate soft-labels $\vy^S = g_S(\vy,\vlambda^{S})$
		\State calculate $\nabla_{\vlambda} \mathcal{L}_\mathrm{v}$ using (\ref{eq:ift2}), where 
		\State $\mathcal{L} = w \mathcal{L}(\vy^S,\hat{\vy})$ and $\mathcal{L}_\mathrm{v} = \mathcal{L}_\mathrm{v}(\vy^\mathrm{v}, \hat{\vy}^\mathrm{v})$ 
		\State update $\vlambda$ using HO optimizer
		\EndFor
		\EndIf
		\EndFor
	\end{algorithmic}
\end{algorithm}

\subsection{AutoDO optimization and complexity}
\label{subsec:complexity}

Our optimization procedure is described in Alg.~\ref{alg:1} with two additional tweaks compared to the theoretical description in Section~\ref{sec:theory}. First, we adopt widely-used stochastic optimization using mini-batches $\mathcal{B}$, which speeds up learning process at the expense of less accurate gradients. Second, the IFT in~(\ref{eq:ift1}) requires to be at a minima point. Hence, we start HO only after reaching epoch $E$ (line 10).

\ifcmt
We estimate complexity of Alg.~\ref{alg:1} in terms of forward and backward passes through a DNN. The conventional inner-objective training (lines 3-9) requires $2N$ passes per epoch. The outer-objective implicit differentiation in lines 12-19 require $(5+T)N$ passes per epoch, where $T$ is the number of Neumann iterations. We set $T=5$ using ablation study in~\cite{lorraine-million} and $E=0.5\times\mathrm{epochs}$. This results in a factor of $(1+0.5(5+T)/2) = 3.5\times$ increase in computation (lines 3-19) compared to only conventional training (lines 3-9). In comparison, DARTS~\cite{liu2018darts} in~\cite{hataya2019faster,li2020dada}, without a finite-difference approximation, has $T=0$ and a factor of $(1+0.5\times5/2) = 2.25\times$ increase in corresponding computation. Runtime estimates are compared in Appendix~A.
\else
We estimate complexity of Alg.~\ref{alg:1} in terms of forward and backward passes through a DNN. The conventional inner-objective training (lines 3-9) requires $2N$ passes per epoch. The outer-objective implicit differentiation in lines 12-19 require $(5+T)N$ passes per epoch, where $T$ is the number of Neumann iterations. We set $T=5$ using ablation study in~\cite{lorraine-million} and $E=0.5\times\mathrm{epochs}$. This results in a factor of $(1+0.5(5+T)/2) = 3.5\times$ increase in computation (lines 3-19) compared to only conventional training (lines 3-9). In comparison, DARTS~\cite{liu2018darts} in~\cite{hataya2019faster,li2020dada}, without a finite-difference approximation, has $T=0$ and a factor of $(1+0.5\times5/2) = 2.25\times$ increase in corresponding computation. Runtime estimates are compared in Appendix~\ref{sec:supp_learning}.
\fi

\section{Experiments}
\label{sec:eval}

\subsection{Experimental setup}
\label{subsec:setup_eval}
We conduct SVHN, CIFAR-10/100 and ImageNet classification experiments. The code is in PyTorch~\cite{paszke2017automatic} with the Kornia library~\cite{eriba2019kornia} for differentiable image operations. We evaluate DA methods using both the near-perfect original train datasets and their distorted versions. To do that, we introduce a \textit{class imbalance} ratio (IR) that scales down the number of available train images for a subset of classes. Class imbalance is defined as the ratio of images with $\{1 \ldots C/2\}$ labels to $\{C/2+1 \ldots C\}$, where $C$ is the total number of classes. Next, we introduce a \textit{label noise} ratio (NR) that defines a proportion of randomly flipped labels in the train dataset. For example, the notation IR-NR=100-0.1 for SVHN in Table~\ref{tab:svhn_results} means that 10\%~(NR=0.1) of random images have randomly flipped labels and the number of images with digits $\{0 \ldots 4\}$ is $100\times$ larger than for $\{5 \ldots 9\}$.

For comparison, there is a variety of gradient-free~\cite{cubuk2018autoaugment,ho2019pba,lim2019fast}, gradient-based~\cite{Lin_2019_ICCV,hataya2019faster,li2020dada} and search-free~\cite{cubuk2019randaugment,lingchen2020uniformaugment} methods described in Section~\ref{sec:related}. We narrow it down to only a single method from each type by excluding~\cite{cubuk2018autoaugment,ho2019pba} due to high complexity and~\cite{Lin_2019_ICCV,hataya2019faster,lingchen2020uniformaugment} due to publicly unavailable code. Hence, we define the following experimental configurations: baseline with standard augmentations, optimization-free RandAugment~\cite{cubuk2019randaugment} (RAA), Bayesian Fast AutoAugment~\cite{lim2019fast} (FAA), gradient-based DADA~\cite{li2020dada}, and our AutoDO. As part of ablation studies, we enable AutoDO features one-by-one with the following notation: $\vlambda^{A_\textrm{SHA}}$ - shared (across all data points) augmentation hyperparameters, $\vlambda^{A}$ - per-point augmentation hyperparameters, $\vlambda^{W}$ - loss reweighting, and $\vlambda^{S}$ - soft-labeling.

\begin{table}[t]
	\caption{WRNet28-10 SVHN top-1 test error rate, $\mu_{\pm\sigma}$ \%.}
	\label{tab:svhn_results}
	\centering
	\begin{tabular}{ccccc}
		\toprule
		Alg./IR-NR& 1-0.0 & 100-0.0 & 1-0.1 & 100-0.1\\
		\midrule
		Baseline                       & 3.6\tiny$\pm$0.10 & 13.6\tiny$\pm$0.69 & 5.3\tiny$\pm$0.27 &20.0\tiny$\pm$1.92\\
		RAA~\cite{cubuk2019randaugment}& 2.7\tiny$\pm$0.04 & 10.9\tiny$\pm$0.66 & 3.4\tiny$\pm$0.11 &13.6\tiny$\pm$0.96\\
		FAA~\cite{lim2019fast}         & 2.8\tiny$\pm$0.02 & 11.5\tiny$\pm$0.32 & 3.7\tiny$\pm$0.08 &15.3\tiny$\pm$1.07\\
		DADA~\cite{li2020dada}         & 2.9\tiny$\pm$0.03 & 12.2\tiny$\pm$0.54 & 4.1\tiny$\pm$0.13 &16.5\tiny$\pm$1.51\\
		\midrule
		$\vlambda^{A_\textrm{SHA}}$ (ours) & 2.8\tiny$\pm$0.10 & 12.6\tiny$\pm$1.53 & 3.0\tiny$\pm$0.17 &13.7\tiny$\pm$0.77\\
		$\vlambda^{A}$              (ours) & 2.7\tiny$\pm$0.09 & 10.2\tiny$\pm$0.50 & 3.0\tiny$\pm$0.07 &12.3\tiny$\pm$0.80\\
		$\vlambda^{A,W}$            (ours) & 2.8\tiny$\pm$0.04 &  6.1\tiny$\pm$0.22 & 2.8\tiny$\pm$0.07 & 8.1\tiny$\pm$0.14\\
		$\vlambda^{A,W,S}$          (ours) & \textbf{2.5}\tiny$\pm$0.04 & \textbf{5.3}\tiny$\pm$0.21 & \textbf{2.6}\tiny$\pm$0.05 & \textbf{6.3}\tiny$\pm$0.57\\
		\bottomrule
	\end{tabular}
\end{table}

\begin{table}[t]
	\caption{WRNet28-10 CIFAR-10 top-1 test error rate, $\mu_{\pm\sigma}$ \%.}
	\label{tab:cifar10_results}
	\centering
	\begin{tabular}{ccccc}
		\toprule
		Alg./IR-NR& 1-0.0 & 10-0.0 & 1-0.1 & 10-0.1\\
		\midrule
		Baseline                       & 6.0\tiny$\pm$0.10 & 16.3\tiny$\pm$0.54 & 12.8\tiny$\pm$0.14 &22.8\tiny$\pm$1.09\\
		RAA~\cite{cubuk2019randaugment}& 5.2\tiny$\pm$0.08 & 13.4\tiny$\pm$0.30 &  7.8\tiny$\pm$0.16 &17.4\tiny$\pm$0.21\\
		FAA~\cite{lim2019fast}         & 5.0\tiny$\pm$0.14 & 13.3\tiny$\pm$0.38 &  8.2\tiny$\pm$0.28 &17.9\tiny$\pm$0.32\\
		DADA~\cite{li2020dada}         & 5.5\tiny$\pm$0.12 & 14.2\tiny$\pm$0.15 & 10.6\tiny$\pm$0.08 &20.2\tiny$\pm$0.83\\
		\midrule
		$\vlambda^{A_\textrm{SHA}}$ (ours) & 5.9\tiny$\pm$0.22 & 17.1\tiny$\pm$0.87 &  7.6\tiny$\pm$0.28 &19.4\tiny$\pm$1.23\\
		$\vlambda^{A}$              (ours) & 5.2\tiny$\pm$0.20 & 13.0\tiny$\pm$0.47 &  7.1\tiny$\pm$0.16 &16.6\tiny$\pm$0.20\\
		$\vlambda^{A,W}$            (ours) & 5.4\tiny$\pm$0.17 & 12.9\tiny$\pm$0.40 &  6.8\tiny$\pm$0.07 &15.1\tiny$\pm$0.19\\
		$\vlambda^{A,W,S}$          (ours) & \textbf{4.9}\tiny$\pm$0.09 & \textbf{11.8}\tiny$\pm$0.35 & \textbf{5.8}\tiny$\pm$0.29 & \textbf{13.1}\tiny$\pm$0.16\\
		\bottomrule
	\end{tabular}
\end{table}

The baseline configuration uses only standard augmentations such as random cropping, horizontal flipping and erasing~\cite{Zhong2020RandomED}. We adopt the publicly available policy models for RAA, FAA and DADA. Using Section~\ref{subsec:augment} format [$\mu$, $\mathrm{rng}$], the list of augmentation operations optimized by AutoDO includes: rotation - $[0^\circ, 30^\circ]$, scale - $[1.0, 0.5]$, translateX/Y - $[0.0, 0.45]$, and shearX/Y - $[0.0, 0.3]$. This list has been selected from the ablation study of the most helpful operations in Cubuk~\etal~\cite{cubuk2019randaugment}. Other operations in AutoDO are applied as in the RAA model.

\ifcmt
We use the same learning hyperparameters for all methods as specified in the Appendix~A. AutoDO optimization starts after the 100th epoch for ImageNet and 50th epoch for others (parameter $E$ in Alg.~\ref{alg:1}). The evaluated architectures are ResNet-18 for ImageNet, Wide-ResNet-28-10 for SVHN and CIFAR-10/100. We run each experiment four times for SVHN and CIFAR-10/100 and report the top-1 mean error rate $\mu$ and standard deviation $\sigma$ on test datasets. We run only one experiment on large-scale ImageNet and report top-1 error rate. Due to a lack of test labels for ImageNet, we use a validation dataset as a test dataset.
\else
We use the same learning hyperparameters for all methods as specified in the Appendix~\ref{sec:supp_learning}. AutoDO optimization starts after the 100th epoch for ImageNet and 50th epoch for others (parameter $E$ in Alg.~\ref{alg:1}). The evaluated architectures are ResNet-18 for ImageNet, Wide-ResNet-28-10 for SVHN and CIFAR-10/100. We run each experiment four times for SVHN and CIFAR-10/100 and report the top-1 mean error rate $\mu$ and standard deviation $\sigma$ on test datasets. We run only one experiment on large-scale ImageNet and report top-1 error rate. Due to a lack of test labels for ImageNet, we use a validation dataset as a test dataset.
\fi

We follow~\cite{lim2019fast} and split original train dataset $\train$ into validation $\validi$ and train $\traini$ datasets using $K$-fold stratified shuffling~\cite{shuffling} for $i$th experiment. The size of $\validi$ is 32\% for SVHN, 20\% for CIFAR-10/100 and ImageNet of the original $\train$. After the split, we distort $\traini$ by introducing class imbalance (IR) and label noise (NR). We estimate AutoDO hyperparameters of $\traini$ using $\validi$, train on $\traini$ and test on $\test$. Unlike~\cite{lim2019fast}, we never train on the original $\train=\traini \cup \validi$ (except for ablations), which leads to lower accuracy compared to ones reported in the corresponding papers due to smaller train dataset size. We choose this approach ($\validi \xrightarrow{\vlambda} \traini$) to obtain undistorted $\validi$ and distorted $\traini$ for our experiments.

\subsection{Quantitative results}
\label{subsec:quant_eval}

Tables~\ref{tab:svhn_results}-\ref{tab:imagenet_results} present test error rates (lower is better) for the selected datasets and model configurations. We clearly see that all DA algorithms outperform the baselines with only standard augmentations. This confirms the common statement that DA improves generalization to test data.

In the case of undistorted train datasets (IR-NR=1-0.0), prior DA algorithms and our AutoDO with augmentations only ($\vlambda^{A}$) tend to perform close to each other, if take into account $\sigma$ error bars. This empirically confirms our proposition in Section~\ref{sec:related} that search-free methods~\cite{cubuk2019randaugment,lingchen2020uniformaugment} are as effective as search-based methods for nearly perfect datasets. The enabled additional AutoDO features along with per-point augmentations ($\vlambda^{A,W,S}$) result in moderate improvements that reach 0.2\% for SVHN, 0.1\% for CIFAR-10 and 0.5\% for CIFAR-100 compared to the best prior methods.

In ablation study, we observe degradation in results by 0.1-1.6\% for AutoDO with the shared hyperparameters ($\vlambda^{A_\textrm{SHA}}$) compared to per-point model ($\vlambda^{A}$). The former configuration is equivalent to a single-policy model, while, for example, a typical RAA~\cite{cubuk2019randaugment}, FAA~\cite{lim2019fast} and DADA~\cite{li2020dada} consist of 25-125 policies. This experimentally verifies the advantage of per-point AutoDO model.

\begin{table}[t]
	\caption{WRNet28-10 CIFAR-100 top-1 test error rate, $\mu_{\pm\sigma}$ \%.}
	\label{tab:cifar100_results}
	\centering
	\begin{tabular}{ccccc}
		\toprule
		Alg./IR-NR& 1-0.0 & 10-0.0 & 1-0.1 & 10-0.1\\
		\midrule
		Baseline                       &24.9\tiny$\pm$0.19 & 45.9\tiny$\pm$0.86 & 36.6\tiny$\pm$0.65 &54.5\tiny$\pm$0.74\\
		RAA~\cite{cubuk2019randaugment}&23.0\tiny$\pm$0.15 & 41.7\tiny$\pm$0.29 & 33.7\tiny$\pm$0.66 &50.1\tiny$\pm$0.52\\
		FAA~\cite{lim2019fast}         &22.7\tiny$\pm$0.23 & 40.0\tiny$\pm$0.27 & 31.4\tiny$\pm$0.25 &48.9\tiny$\pm$0.51\\
		DADA~\cite{li2020dada}         &23.1\tiny$\pm$0.05 & 41.1\tiny$\pm$0.22 & 32.4\tiny$\pm$0.48 &49.6\tiny$\pm$0.44\\
		\midrule
		$\vlambda^{A_\textrm{SHA}}$ (ours) & 24.7\tiny$\pm$0.94 & 41.0\tiny$\pm$1.29 & 31.4\tiny$\pm$1.75 & 48.9\tiny$\pm$1.80\\
		$\vlambda^{A}$              (ours) & 23.1\tiny$\pm$0.25 & 40.8\tiny$\pm$0.43 & 30.6\tiny$\pm$0.48 & 47.4\tiny$\pm$0.33\\
		$\vlambda^{A,W}$            (ours) & \textbf{22.2}\tiny$\pm$0.22 & 39.0\tiny$\pm$0.30 & 30.6\tiny$\pm$0.50 & 48.0\tiny$\pm$0.38\\
		$\vlambda^{A,W,S}$          (ours) & 22.4\tiny$\pm$0.25 & \textbf{36.7}\tiny$\pm$0.32 & \textbf{27.3}\tiny$\pm$0.29 & \textbf{39.6}\tiny$\pm$0.06\\
		\bottomrule
	\end{tabular}
\end{table}

\begin{table}[t]
	\caption{ResNet18 ImageNet top-1 test error rate, \%.}
	\label{tab:imagenet_results}
	\centering
	\begin{tabular}{ccc}
		\toprule
		Alg./IR-NR& 10-0.0 & 10-0.1\\
		\midrule
		Baseline                       & 43.1 & 46.2\\
		RAA~\cite{cubuk2019randaugment}& 42.0 & 45.4\\
		FAA~\cite{lim2019fast}         & 41.6 & 44.7\\
		DADA~\cite{li2020dada}         & 41.9 & 44.9\\
		\midrule
		$\vlambda^{A,W,S}$          (ours) & \textbf{40.4} & \textbf{44.1}\\
		\bottomrule
	\end{tabular}
\end{table}

Next, we compare robustness of all methods to train dataset distortions. In the case of class imbalance only (IR=10/100, NR=0.0), AutoDO achieves at least 5.6\%, 1.5\%, 3.3\% and 1.2\% gain compared to prior methods for SVHN, CIFAR-10/100, and ImageNet, respectively. For instance, the AutoDO loss reweighting sub-model decreases SVHN error rate from 10.2\% for our model with $\vlambda^{A}$ to 6.1\% for the model with $\vlambda^{A,W}$. When we introduce 10\% labels noise only (IR-NR=1-0.1), our improvements are, correspondingly, 0.8\%, 2.0\% and 4.1\% for SVHN and CIFAR-10/100 compared to the best prior methods. It is mostly achieved by enabling the soft-label estimation sub-model. For example, CIFAR-100 test error rate decreased from 30.6\% for our model with $\vlambda^{A,W}$ to 29.4\% for $\vlambda^{A,W,S}$.

When we distort train data by both class imbalance and label noise (IR=10/100, NR=0.1), AutoDO achieves at least 7.3\%, 4.3\%, 9.3\% and 0.6\% gain for SVHN, CIFAR-10/100, and ImageNet, respectively. To sum up, all components of our model gradually improve classification accuracy for all datasets. The only exception is CIFAR-100, where the soft-label sub-model degrades mean accuracy by 0.2\% in the case of undistorted train dataset. This can be related to suboptimal initialization constant $\alpha=0.1$ for 100 classes. Also, we notice that the gain on ImageNet is lower (0.6-1.2\%) compared to small-scale datasets. The latter might be related to a relatively shallow ResNet18 model, which has been chosen to decrease experiment time.

\begin{figure}[t]
	\centering
	\includegraphics[width=0.98\columnwidth]{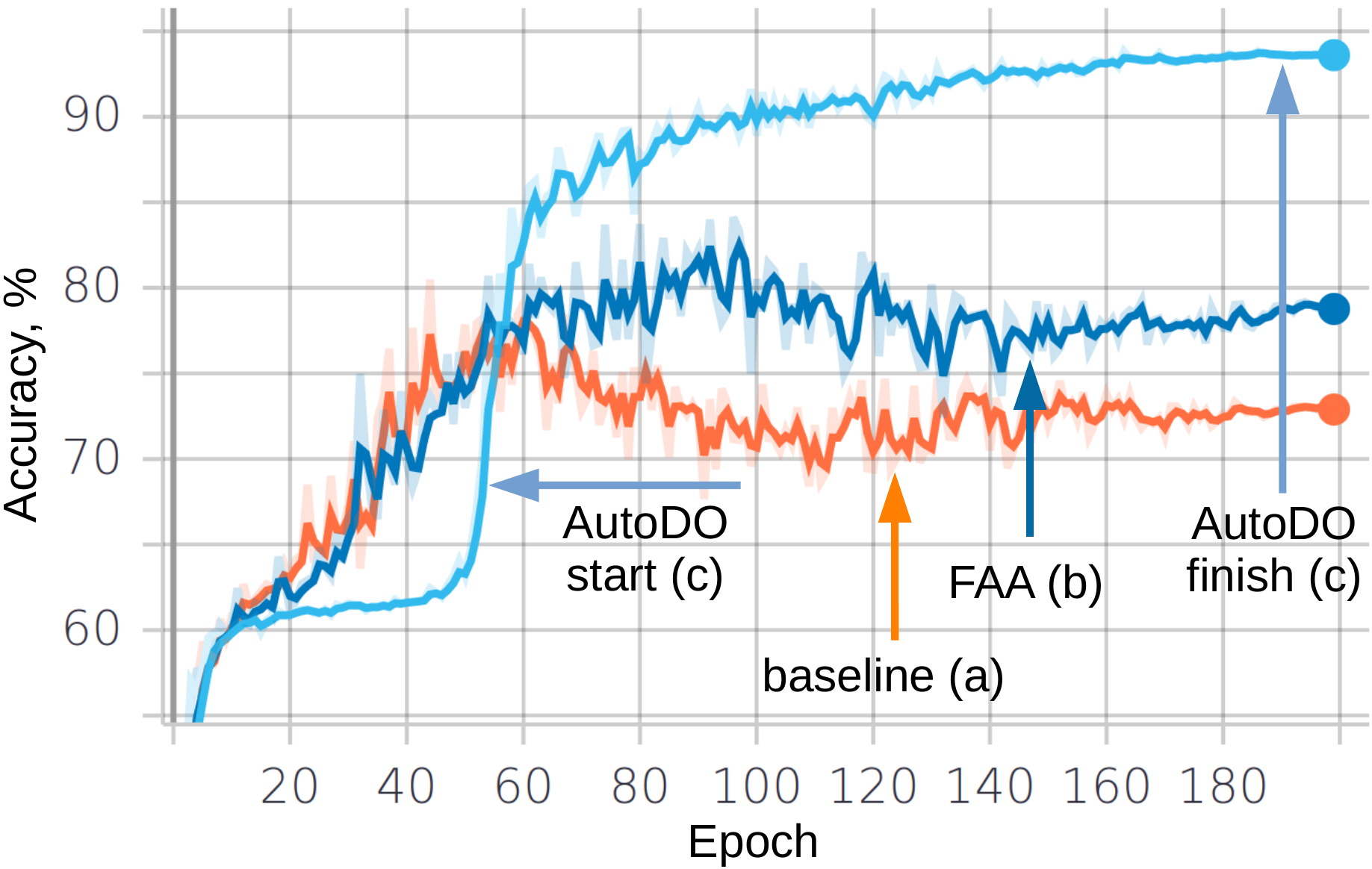}
	\caption{SVHN top-1 test accuracy learning curve for the models trained on a dataset with 100$\times$ class imbalance and 10\% label noise: (a) baseline, (b) FAA~\cite{lim2019fast}, and (c) our AutoDO ($\vlambda^{A,W,S}$). Our optimization starts at epoch $E=50$. It prevents overfitting to the distorted train data and improves generalization to test data.}
	\label{fig:svhn_curve}
\end{figure}

\ifcmt
Figure~\ref{fig:svhn_curve} illustrates the typical learning process on distorted train data, specifically, for SVHN with $100\times$ class-imbalance and 10\% label noise. It is evident that FAA~\cite{lim2019fast} (b) improves accuracy compared to the baseline (a), but it quickly starts to overfit to noisy labels~\cite{zhangoverfit} and over-represented classes. The AutoDO optimization starts at epoch $E=50$ and avoids overfitting to the distorted train data by selecting proper per-point hyperparameter vectors $\vlambda^{A,W,S}$. Hence, we conclude that AutoDO is more robust to distorted data. Learning curves for train and test losses are given in the Appendix~B.
\else
Figure~\ref{fig:svhn_curve} illustrates the typical learning process on distorted train data, specifically, for SVHN with $100\times$ class-imbalance and 10\% label noise. It is evident that FAA~\cite{lim2019fast} (b) improves accuracy compared to the baseline (a), but it quickly starts to overfit to noisy labels~\cite{zhangoverfit} and over-represented classes. The AutoDO optimization starts at epoch $E=50$ and avoids overfitting to the distorted train data by selecting proper per-point hyperparameter vectors $\vlambda^{A,W,S}$. Hence, we conclude that AutoDO is more robust to distorted data. Learning curves for train and test losses are given in the Appendix~\ref{sec:supp_curves}.
\fi

\begin{table}[t]
	\caption{AutoDO on $\validi$ vs. $\test$, test error rate, $\mu_{\pm\sigma}$ \%.}
	\label{tab:test_overfit}
	\centering
	\begin{tabular}{ccc}
		\toprule
		Dataset/Split & $\validi \xrightarrow{\vlambda} \traini$ & $\test \xrightarrow{\vlambda} \traini$\\
		\midrule
		SVHN      & 2.48\tiny$\pm$0.04 & 2.48\tiny$\pm$0.05 \\
		CIFAR-10  & 4.92\tiny$\pm$0.09 & 5.11\tiny$\pm$0.12 \\
		CIFAR-100 &22.20\tiny$\pm$0.22 &22.12\tiny$\pm$0.30 \\
		\bottomrule
	\end{tabular}
\end{table}

\textbf{Ablation study: overfitting to test data}. In our data setup in Figure~\ref{fig:general}, we assume that small validation dataset approximates test data distribution $Q^{\mathrm{test}}_{\vx,\vy}$. To verify this and to check how well the learned hyperparameters generalize to unseen test data, we replace $\validi$ with $\test$ in AutoDO estimation step (line 13 in Alg.~\ref{alg:1}). Table~\ref{tab:test_overfit} compares AutoDO results in both setups. Due to minor differences in error rates, we conclude that, indeed, $Q^{\mathrm{test}}_{\vx,\vy} \approx Q^{\mathrm{val}}_{\vx,\vy}$ and no significant overfitting to $\validi$ happens.

\begin{table}[t]
	\caption{Split as in~\cite{lim2019fast}: $\validi \xrightarrow{\vlambda} \train$, test error rate, $\mu_{\pm\sigma}$ \%.}
	\label{tab:test_oversplit}
	\centering
	\begin{tabular}{cccc}
		\toprule
		Dataset/Alg. & RAA~\cite{cubuk2019randaugment} & FAA~\cite{lim2019fast} & AutoDO\\
		\midrule
		SVHN      & 2.46\tiny$\pm$0.05 & 2.49\tiny$\pm$0.05 & 2.39\tiny$\pm$0.02 \\
		CIFAR-10  & 4.70\tiny$\pm$0.13 & 4.37\tiny$\pm$0.11 & 4.56\tiny$\pm$0.16 \\
		CIFAR-100 &21.85\tiny$\pm$0.21 &20.42\tiny$\pm$0.21 &20.49\tiny$\pm$0.27 \\
		\bottomrule
	\end{tabular}
\end{table}

\textbf{Ablation study: training and estimating $\vlambda$ on all train data}: $\validi \xrightarrow{\vlambda} \train$, where $\train = \traini \cup \validi$. We conduct experiments for the Lim~\etal~\cite{lim2019fast} data split: AutoDO dataset hyperparameters are estimated from $\validi$ and applied to the original $\train$. This increases train dataset size compared to experiments in Tables~\ref{tab:svhn_results}-\ref{tab:cifar100_results}. Unlike~\cite{lim2019fast} with explicit steps for policy merge after each data split step, we randomly sample mini-batches for $\traini$ and $\validi$ from $\train$ in lines 12-13 of Alg.~\ref{alg:1}. Table~\ref{tab:test_oversplit} contains results for the original train datasets. Although error rates decrease by 0.1\% for SVHN, 0.3\% for CIFAR-10 and 1.9\% for CIFAR-100 due to larger train datasets, there is no significant difference in results between the best prior methods and AutoDO for nearly perfect train data.

\begin{figure*}[t]
	\centering
	\includegraphics[width=0.71\textwidth]{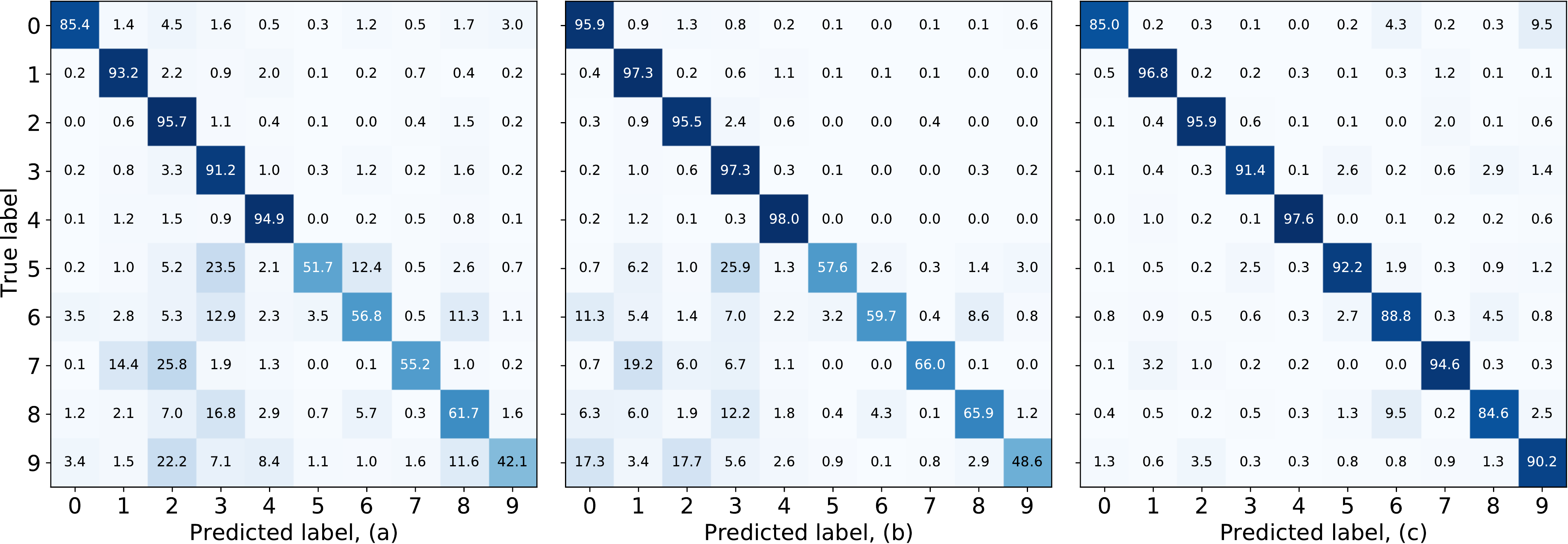}
	\includegraphics[width=0.71\textwidth]{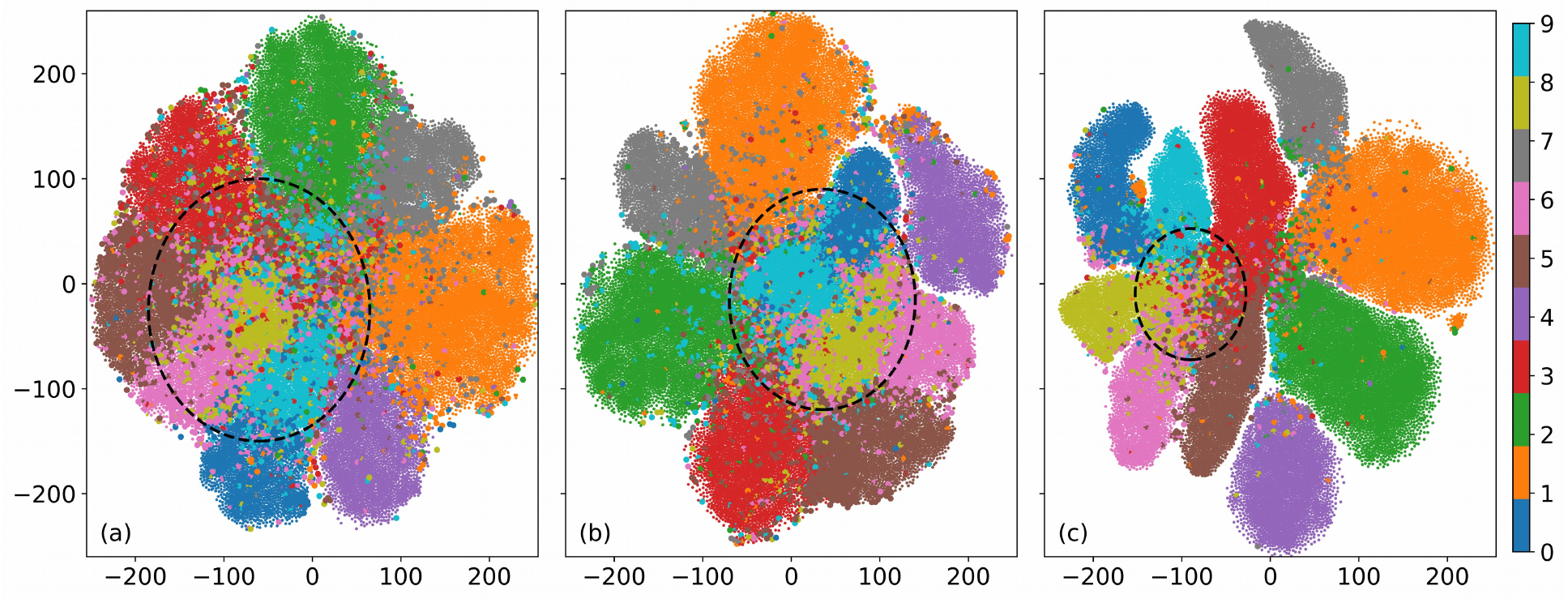}
	\caption{Confusion matrices (top) of SVHN test data and t-SNE clusters of penultimate layer embeddings (bottom) for the models trained on a dataset with 100$\times$ class imbalance and 10\% label noise: (a) baseline, (b) FAA~\cite{lim2019fast}, and (c) our AutoDO ($\vlambda^{A,W,S}$). Dots and balls represent correctly and incorrectly classified images for t-SNE visualizations, respectively. The underrepresented classes $\{5\ldots9\}$ have on average 53.5\% accuracy for baseline~(a), 59.6\% for FAA~\cite{lim2019fast}~(b), and 90.1\% for our method~(c).}
	\label{fig:svhn_cmat}
\end{figure*}

\subsection{Qualitative results}
\label{subsec:qual_eval}

We present additional evidence that our AutoDO model solves the shared-policy DA limitations sketched in Figure~\ref{fig:prob}. To do that, we select three models from Table~\ref{tab:svhn_results} for SVHN: (a) baseline with standard augmentations only, (b) shared-policy FAA~\cite{lim2019fast}, and (c) our per-point AutoDO model with all features ($\vlambda^{A,W,S}$). These models have been trained on a distorted dataset with 100$\times$ class imbalance and 10\% label noise, and tested on $\test$.

First, we calculate confusion matrices of test data for each trained model. Figure~\ref{fig:svhn_cmat}~(top) shows that the shared-policy FAA~\cite{lim2019fast} model improves the average top-1 accuracy compared to the baseline (from 78.4\% to 83.1\%), but the improvement for the class-imbalanced digits $\{5\ldots9\}$ (from 53.5\% to 59.6\%) is approximately the same as the average gain. In contrast, our AutoDO model aligns the distribution of the underrepresented digits with the unbiased test data and their accuracy improves from 53.5\% to 90.1\% with average accuracy of 93.0\%. The standard deviation of the inter-class accuracy drops from 20\% for (a) and 19.2\% for (b) to only 4.4\% for our AutoDO model (c).

Second, we cluster test data embeddings of the trained models using the popular t-SNE~\cite{tsne} method. These embeddings are the output of WRNet28-10~\cite{Zagoruyko2016WRN} penultimate layer that precedes a linear classifier. This setup exactly matches Figure~\ref{fig:prob} setting. T-SNE clusters are shown in Figure~\ref{fig:svhn_cmat}~(bottom), where dots and balls represent correctly and incorrectly classified images, respectively. While FAA~\cite{lim2019fast} (b) has significantly less misclassified digits (number of balls) compared to the baseline (a), it is unable to clearly separate clusters of underrepresented digits \eg "6", "8" and "9". These clusters are heavily mixed due to the limitations of shared-policy DA model. Our AutoDO model solves this issue by per-point augmentation and loss reweighting: "6", "8" and "9" clusters are much better separated, which explains 36.6\% increase in accuracy. Furthermore, our soft-label estimation allows to enlarge margins between not only clusters of underrepresented digits $\{5\ldots9\}$, but also well-represented digits $\{0\ldots4\}$.

\section{Conclusions}
\label{sec:conclusion}

We introduced the AutoDO model that addresses the limitations of existing policy-based AutoAugment models by incorporating the set of hyperparameters independently for each train data point. We jointly optimized these large-scale hyperparameters using an elegant low-complexity implicit differentiation framework. Our probabilistic interpretation using Fisher information matched the objective of minimizing the distribution shift between small noise-free unbiased validation dataset and large distorted train dataset.

AutoDO components such as augmentation hyperparameters, loss weights and soft-labels showed their effectiveness on SVHN, CIFAR-10/100 and ImageNet classification in ablation studies. For class-imbalanced train datasets with label noise, our AutoDO improved the average accuracy and, more importantly, aligned the precision on underrepresented classes by a significant margin compared to previous methods. Hence, our AutoDO is more robust in real applications with imperfect train datasets.

{\small
  \bibliographystyle{ieee_fullname}
  \bibliography{05036}
}

\ifcmt
  \end{document}
\else
  \appendix
\section{Experiments: learning hyperparameters}
\label{sec:supp_learning}

We train task model with the following hyperparameters for all methods and datasets: SGD optimizer with 0.9 Nesterov momentum, 1e-4 weight decay, 200 train epochs, 256 mini-batch size, cosine learning rate annealing with 5 warm-up epochs. Learning rate is 0.005 for SVHN and 0.1 for CIFAR-10/100 and ImageNet. The HO optimizer is RMSprop with 0.01 learning rate for ImageNet and 0.05 for others, and it uses the same schedule as the task optimizer.

\begin{table}[h]
	\caption{Runtime comparison of our and recently published methods, GPU hours. AutoDO results are estimated using V100 GPU.}
	\label{tab:supp_runtime}
	\centering
	\begin{tabular}{ccc}
		\toprule
		Alg./Dataset& CIFAR-10 & SVHN\\
		\midrule
		AA~\cite{cubuk2018autoaugment} & 5,000 & 1,000\\
		PBA~\cite{ho2019pba}           & 5.0 & 1.0\\
		FAA~\cite{lim2019fast}         & 3.5 & 1.5\\
		OHL~\cite{Lin_2019_ICCV}       & 83.3& -\\
		DADA~\cite{li2020dada}         & 0.1 & 0.1\\
		\midrule
		AutoDO, 1 epoch         & 0.036& 0.046\\
		AutoDO, $E=150$         & 1.8  & 2.3\\
		AutoDO, $E=50$          & 5.4  & 6.8\\
		\bottomrule
	\end{tabular}
\end{table}

We use only $\vtheta$ of penultimate fully-connected layer in~(\ref{eq:ift2}) with $T=5$ Neumann series iterations for $\mH_{\vtheta}^{-1}$ approximation. The $T$ hyperparameter is chosen from Fig. 3 ablation study in~\cite{lorraine-million}. The hyperparameter $E$ in Alg.~1 is $100$ for ImageNet and $50$ for others. The soft-label initialization constant from Section 4.2 is $\alpha=0.1$. The evaluated architectures are ResNet-18 for ImageNet, Wide-ResNet-28-10 for CIFAR-10/100 and SVHN. We replace ReLU nonlinearities in all networks with CELU~\cite{celu} to satisfy $C^1$ requirement in~(\ref{eq:ift1}). While IFT in~(\ref{eq:ift1}) is only locally defined $(\| \vlambda - \acute{\vlambda} \| \le r_1$ and $\| \vtheta - \acute{\vtheta} \| \le r_2)$ for a fixed point $(\acute{\vlambda},\acute{\vtheta})$, practically, the choice of hyperparameter $E$ and CELU stabilizes optimization process in Alg.~\ref{alg:1}. In addition, there are an attempts to extend IFT to a global case in~\cite{idczak, cristea, galewski2017global}.

\begin{figure}[ht]
	\centering
	\includegraphics[width=0.80\columnwidth]{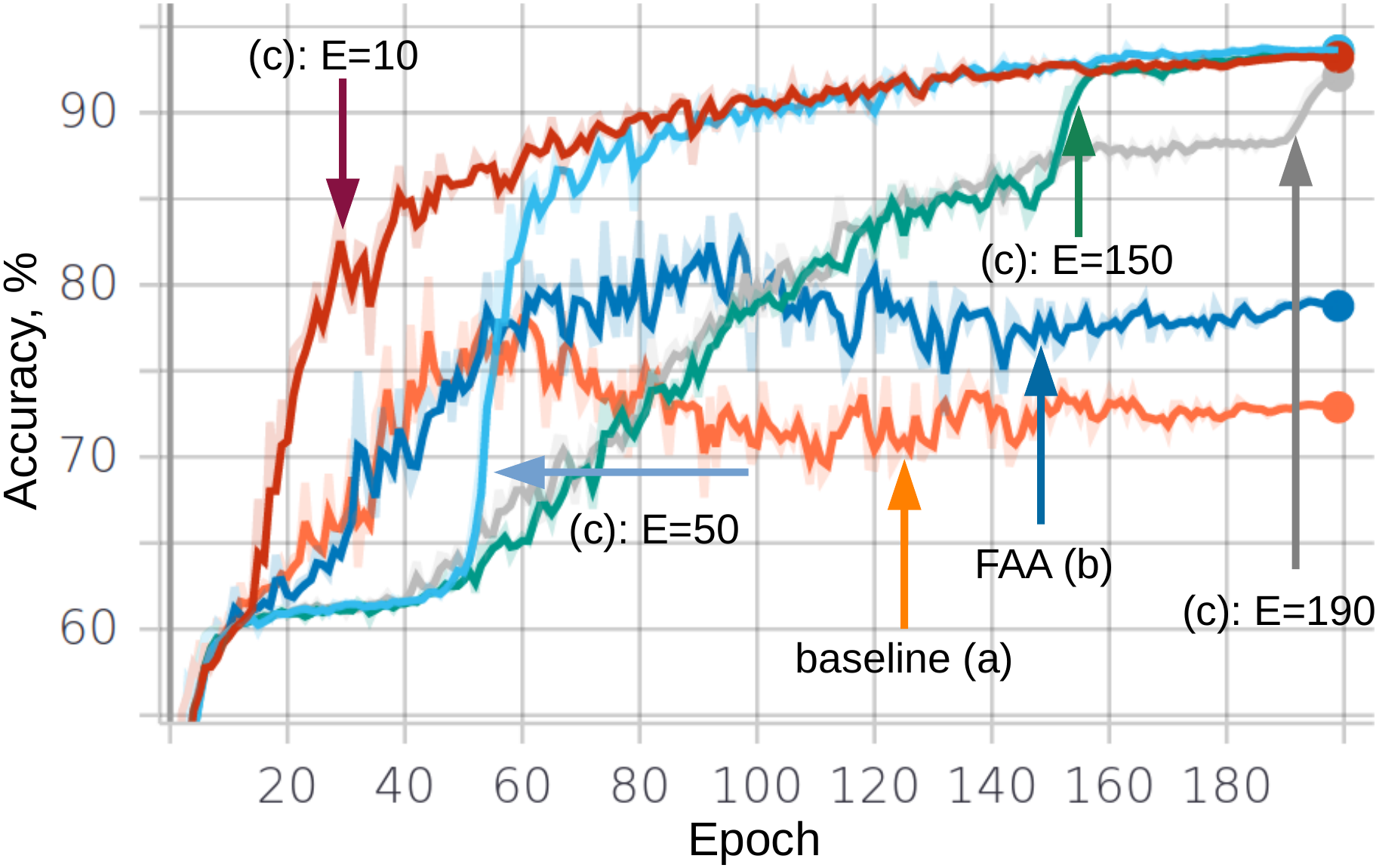}
	\includegraphics[width=0.80\columnwidth]{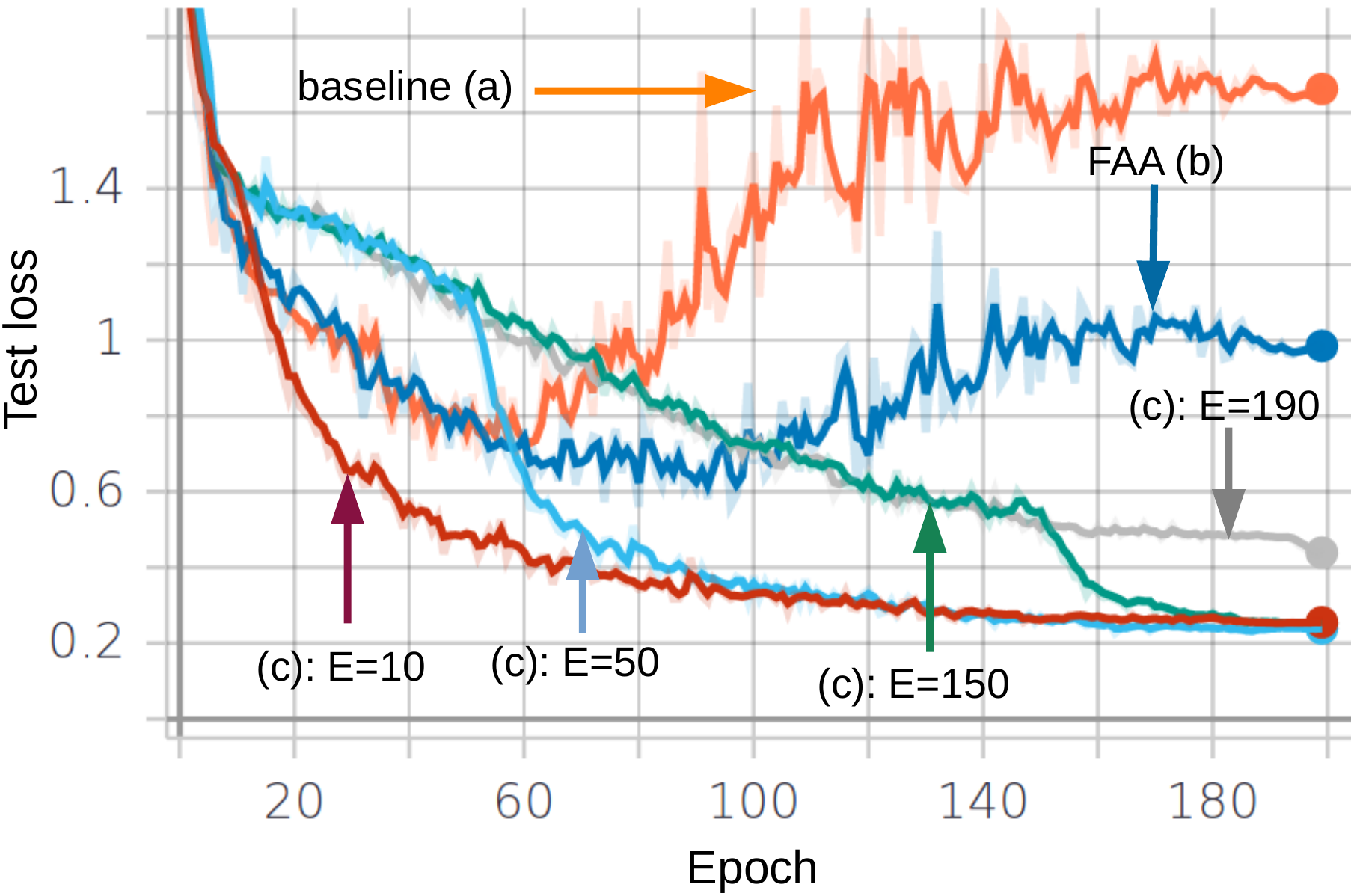}
	\includegraphics[width=0.80\columnwidth]{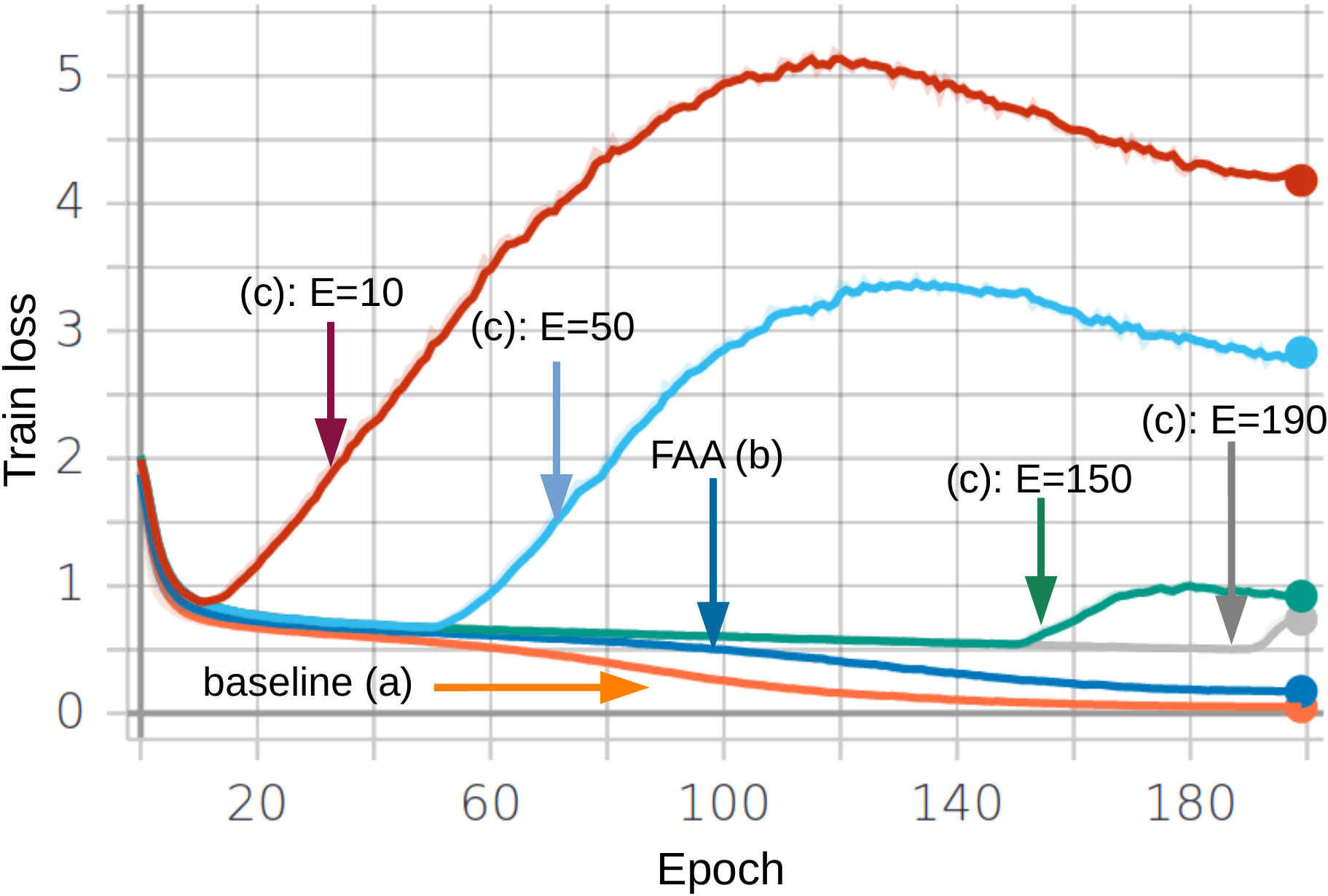}
	\caption{SVHN top-1 test accuracy (top), test loss (middle) and train loss (bottom) learning curves for the following models trained on a dataset with 100$\times$ class imbalance and 10\% label noise: (a) baseline, (b) FAA~\cite{lim2019fast}, and (c) our AutoDO with $\vlambda^{A,W,S}$. Our HO starts at epoch $E=\{10,50,150,190\}$, which prevents overfitting to the distorted train data and improves generalization to test data.}
	\label{fig:supp_svhn_curve}
\end{figure}

We run all experiments on P100/V100 GPUs and each one takes few hours depending on the dataset setting. Only large-scale ImageNet experiments can take up to several days. Table~\ref{tab:supp_runtime} presents detailed comparison between the estimated AutoDO runtime and the reported runtimes for other recent methods.

\section{Experiments: SVHN learning curves}
\label{sec:supp_curves}

The additional learning curves for SVHN are showed in Figure~\ref{fig:supp_svhn_curve}. Our AutoDO optimization starts at epoch $E=\{10,50,150,190\}$. It is evident that $E=50$ hyperparameter has a minor edge over other settings, but practically $E=150$ works almost the same with $3\times$ less computing.

\section{Theoretical background: implicit differentiation meets density matching}
\label{sec:supp_density}

The gradient of expectation in~(\ref{eq:meet3}) over two independent distributions $\hat{Q}^{\mathrm{val}}_{\vx}$ and $\hat{Q}_{\vx}$ is a product
\begin{equation*} \label{eq:supp-meet1}
\nabla_{\vlambda} \sE_{\hat{Q}^\mathrm{val}_{\vx}, \hat{Q}_{\vx}} \left[ \mathcal{L}_\mathrm{v} \right] = - \sE_{\hat{Q}^\mathrm{val}_{\vx}} \left[ \nabla_{\vtheta} \mathcal{L}_\mathrm{v} \right] \sE_{\hat{Q}_{\vx}} \left[ \mH_{\vtheta}^{-1} \nabla_{\vtheta} \nabla_{\vlambda}^T \mathcal{L} \right].
\end{equation*}

The Hessian $\mH_{\vtheta}$ itself is calculated as expectation over $\hat{Q}_{\vx}$ due to computational and stability reasons. It leads to a connection to the Fisher information metric $\mathcal{\mI}_{\vtheta}$~\cite{lyfisher} as
\begin{equation*} \label{eq:supp-meet2}
\sE_{\hat{Q}_{\vx}} \left[ \mH_{\vtheta} \right] = \sE_{\hat{Q}_{\vx}} \left[ \frac{\partial^2 \mathcal{L}}{\partial \vtheta \partial \vtheta^T} \right] = \sE_{\hat{Q}_{\vx}} \left[ \vu_i(\vtheta) \vu_i(\vtheta)^T \right] = -\mathcal{\mI}_{\vtheta},
\end{equation*}
where $\vu_i(\vtheta)=-\partial \mathcal{L}(i)/\partial \vtheta=\nabla_{\vtheta}\log p(\vy_i|\vx_i, \vtheta(\vlambda))$ are the Fisher scores~\cite{learningfisher} for $\log$-likelihood loss function.

The expectation $\sE_{\hat{Q}_{\vx}} \left[ \nabla_{\vtheta} \nabla_{\vlambda}^T \mathcal{L} \right]$ contains a second-order derivative of the train loss $\mathcal{L}(i) = -\log p(\vy_i|\vx_i, \vtheta(\vlambda)) = -\log p$ from~(6) that can be similarly simplified to
\begin{equation*} \label{eq:supp-meet3}
\begin{split}
&\sE_{\hat{Q}_{\vx}} \left[ \frac{\partial^2\mathcal{L}}{\partial\vtheta \partial\vlambda^T} \right] = -\sE_{\hat{Q}_{\vx}} \left[ \frac{\partial}{\partial\vtheta} \left( \frac{\partial \log p}{\partial\vlambda^T} \right) \right] = \\
&-\sE_{\hat{Q}_{\vx}} \left[ -\frac{1}{p^2} \frac{\partial p}{\partial\vtheta} \frac{\partial p}{\partial\vlambda^T} + \frac{1}{p} \frac{\partial^2 p}{\partial\vtheta \partial\vlambda^T} \right] =
\sE_{\hat{Q}_{\vx}} \left[ \frac{\partial \mathcal{L}}{\partial\vtheta} \frac{\partial \mathcal{L}}{\partial\vlambda^T}\right],
\end{split}
\end{equation*}
where the term $-\sE_{\hat{Q}_{\vx}} \left[\frac{1}{p} \frac{\partial^2 p}{\partial\vtheta \partial\vlambda^T} \right] = -\frac{\partial^2}{\partial\vtheta \partial\vlambda^T} \int_x p d\vx = 0$.

By substituting the above derivations, the final form of~(\ref{eq:meet3}) and its practical variant with equal-probability data points  can be obtained as
\begin{equation*} \label{eq:supp-meet4}
\begin{split}
&\nabla_{\vlambda} \sE_{\hat{Q}^\mathrm{val}_{\vx}, \hat{Q}_{\vx}} \left[ \mathcal{L}_\mathrm{v} \right] = \sE_{\hat{Q}^\mathrm{val}_{\vx}} \left[ \vu^\mathrm{v}(\vtheta) \right] \mathcal{\mI}_{\vtheta}^{-1} \sE_{\hat{Q}_{\vx}} \left[ \vu(\vtheta) \vu(\vlambda)^T \right]\\
& = \left[ \frac{1}{M} \sum_{j \in \sM} \vu^\mathrm{v}_j(\vtheta) \right] \mathcal{\mI}_{\vtheta}^{-1} \left[ \frac{1}{N} \sum_{i \in \sN} \vu_i(\vtheta) \vu^T_i(\vlambda) \right].
\end{split}
\end{equation*}

  \end{document}
\fi